\documentclass[sigconf]{acmart}
\AtBeginDocument{%
  }

\setcopyright{acmlicensed}
\copyrightyear{2018}
\acmYear{2018}
\acmDOI{XXXXXXX.XXXXXXX}
\acmConference[Conference acronym 'XX]{Make sure to enter the correct
  conference title from your rights confirmation email}{June 03--05,
  2018}{Woodstock, NY}
\acmISBN{978-1-4503-XXXX-X/2018/06}

\begin{document}

\title{Beyond Retrieval-Ranking: A Multi-Agent Cognitive Decision Framework for E-Commerce Search}

\author{Zhouwei Zhai}
\authornote{The corresponding author.}
\orcid{1234-5678-9012}
\email{zhaizhouwei1@jd.com}
\affiliation{%
  \institution{JD.com}
  \city{Beijing}
  \country{China}
}

\author{Mengxiang Chen}
\email{chenmengxiang9@jd.com}
\affiliation{%
  \institution{JD.com}
  \city{Beijing}
  \country{China}
}
\author{Haoyun Xia}
\email{xiahaoyun1@jd.com}
\affiliation{%
  \institution{JD.com}
  \city{Beijing}
  \country{China}
}
\author{Jin Li}
\email{lijin.257@jd.com}
\affiliation{%
  \institution{JD.com}
  \city{Beijing}
  \country{China}
}
\author{Renquan Zhou}
\email{zhourenquan.1@jd.com}
\affiliation{%
  \institution{JD.com}
  \city{Beijing}
  \country{China}
}
\author{Min Yang}
\email{yangmin.aurora@jd.com}
\affiliation{%
  \institution{JD.com}
  \city{Beijing}
  \country{China}
}

\renewcommand{\shortauthors}{Zhouwei et al.}

\begin{abstract}
The retrieval-ranking paradigm has long dominated e-commerce search, but its reliance on query-item matching fundamentally misaligns with multi-stage cognitive decision processes of platform users. This misalignment introduces critical limitations: semantic gaps in complex queries, high decision costs due to cross-platform information foraging, and the absence of professional shopping guidance. To address these issues, we propose a Multi-Agent Cognitive Decision Framework (MACDF), which shifts the paradigm from passive retrieval to proactive decision support. Extensive offline evaluations demonstrate MACDF’s significant improvements in recommendation accuracy and user satisfaction, particularly for complex queries involving negation, multi-constraint, or reasoning demands. Online A/B testing on JD search platform confirms its practical efficacy. This work highlights the transformative potential of multi-agent cognitive systems in redefining e-commerce search.
\end{abstract}




\maketitle

\section{Introduction}
The retrieval-ranking paradigm has served as the cornerstone of e-commerce search for decades. Its evolution has primarily centered on optimizing increasingly complex retrieval models\cite{qu2020rocketqa,magnani2022semantic,peng2023entity,nigam2019semantic,wang2023learning} and ranking models\cite{liu2022knowledge,zhou2018deep,yan2018beyond} to enhance click-through rate (CTR) and conversion rate (CVR). With the rapid advancement of large language models (LLMs), LLM-based optimization schemes for e-commerce search are now being extensively explored\cite{peng2024large,li2025matching,asai2024self,sun2023learning,dai2024enhancing,thomas2024large}. However, these optimizations remain confined within the retrieval-ranking paradigm, which is fundamentally misaligned with users' cognitive decision-making processes during shopping. We observe that for many queries, particularly those with complex intents, systems often fail to adequately meet user needs, necessitating multiple rounds of active query modification by users. Furthermore, the journey from initiating a search to finalizing a purchase typically involves numerous clicks to view product details and user reviews, often with repeated visits, and may even include cross-platform information-seeking behavior. These phenomena reveal a profound architectural flaw: the simplification of purchasing behavior into mere query-item matching overlooks the user's multi-stage cognitive process of gathering, evaluating, and integrating fragmented information.

This misalignment gives rise to three major systemic deficiencies. First, there exists a semantic gap in complex scenarios: when a user searches for a "lightweight laptop suitable for 4K video editing", keyword-based models might only match tags like "laptop" and "4K", failing to infer implicit requirements such as "VRAM capacity" or "cooling performance", thereby forcing users into tedious query reformulation. Second, a high decision cost is imposed: users are
compelled to piece together fragmented information across multiple
platforms (product detail pages, review communities, knowledge forums),
a cost that is entirely invisible in conventional CTR/CVR metrics. Third, there is an absence of expert guidance: current systems only provide shallow signals(e.g., product information, reviews, ratings, sales volume) but cannot answer expert-level questions such as "Can this solid-state drive withstand daily video rendering workloads?".

Addressing these challenges necessitates a paradigm shift from passive product matching to proactive support for shopping decisions. To this end, we propose a Multi-Agent Cognitive Decision Framework (MACDF). This architecture replaces the bottleneck stages dominated by retrieval-ranking in traditional e-commerce search with a collaborative ensemble of specialized agents, simulating the role of professional shopping consultants. Its core innovation lies in redefining the optimization objective around minimizing the user's decision-making cost, thereby dynamically optimizing the purchase path while providing the specialized decision guidance long absent in traditional e-commerce systems.

The contributions of this work are as follows: (1) We are the first to reframe e-commerce search as a Multi-Agent Cognitive Decision Framework (MACDF), moving beyond the conventional retrieval-ranking paradigm. (2) The design of MACDF incorporates various specialized agents (e.g., Leader, Guider, Planner, ProductSearch, WebSearch, and Decider) that learn collaborative strategies to maximize decision-making efficiency and final conversion rates.

To validate its effectiveness, we conducted extensive offline evaluations on an E-commerce Cognitive Decision Benchmark (ECCD-Bench) constructed from real-world e-commerce search logs. The results demonstrate that MACDF significantly improves both product recommendation accuracy and user demand satisfaction (UDS), particularly for complex query intents involving negations, multiple constraints, and reasoning. Online A/B testing conducted on the JD platform confirmed the practical efficacy of MACDF. Compared to the existing production system, MACDF achieved a relative improvement of 6.5\% in UCVR, a relative improvement of 3.9\% in UCTR, and a relative reduction of 7.7\% in the Average Reformulation Count (ARC). These results substantiate that MACDF successfully addresses the core cognitive challenges in e-commerce search, enhancing both user experience and key business metrics.

\section{Related Work}
\subsection{Traditional E-commerce Search}
Traditional e-commerce search systems are built upon and optimized within the two-stage retrieval-ranking paradigm. Classical methods for retrieval, such as LSI\cite{deerwester1990indexing}, BM25\cite{robertson2009probabilistic}, and TF-IDF\cite{aizawa2003information}, primarily rely on statistical features. Embedding approaches like RocketQA\cite{qu2020rocketqa}, DPR\cite{karpukhin2020dense} and\cite{Li2021EmbeddingbasedPR}\cite{Kekuda2024EmbeddingBR}\cite{He2023Que2EngageER} aim to address the lexical gap in keyword matching by learning query-product vector representations. In ranking optimization, knowledge distillation\cite{liu2022knowledge} has been employed to refine product ordering. Personalized ranking has been enhanced through methods like DIN\cite{zhou2018deep} and personalized vector modeling\cite{yan2018beyond}, while step-by-step closed-loop optimization has also been explored\cite{subramaniam2025ai}. Major e-commerce platforms continuously investigate advanced neural network techniques\cite{magnani2022semantic,peng2023entity,nigam2019semantic,wang2023learning}. However, these methods often lack compositional reasoning capabilities for handling strong semantic constraints (e.g., "100-megapixel phone but not Xiaomi") or multi-attribute implicit needs (e.g., "anti-aging serum under 100 RMB suitable for oily, sensitive skin"), leading to a pronounced semantic gap. The retrieval stage is further hampered by static indexing, which prevents the dynamic decomposition of complex query intents (e.g., "anti-aging" + "moisturizing" + "for sensitive skin"), resulting in poor coverage for long-tail queries.

\subsection{LLM-enhanced Search}

The rapid advancement of Large Language Models (LLMs) has significantly propelled the optimization of search systems. This includes LLM-based query rewriting and expansion\cite{peng2024large} to improve product recall, generative retrieval\cite{li2025matching,asai2024self,sun2023learning}, fine-tuning LLMs to replace or enhance traditional relevance models for better semantic matching accuracy\cite{dai2024enhancing}, and utilizing LLMs for relevance annotation\cite{thomas2024large}. Nevertheless, these LLM-enhanced optimizations remain confined within the conventional retrieval-ranking framework. They fail to cover the entire shopping decision-making pipeline and lack mechanisms for synergistic optimization across different stages. Furthermore, they do not provide structured decision support capabilities (e.g., product comparison matrices, professional purchasing advice), thus failing to effectively reduce the user's decision cost during shopping.

\subsection{Multi-Agent Systems}
General-purpose multi-agent frameworks\cite{li2024survey} demonstrate effective collaborative problem-solving and are often viewed as a path toward superhuman intelligence. Frameworks like ReAct\cite{yao2023react} integrate reasoning and acting, enhancing agent robustness through reflective loops. MetaGPT\cite{hong2024metagpt} improves multi-agent collaboration by standardizing workflows and generating structured intermediate outputs (e.g., requirement documents, API call chains). However, these multi-agent systems are not specifically designed or optimized for e-commerce decision-making scenarios. They lack the injection of domain-specific e-commerce knowledge and do not model critical factors such as inter-attribute relationships or price sensitivity, leading to recommendations that are often impractical for real-world purchasing decisions.

\section{Framework}

This section introduces the Multi-Agent Cognitive Decision Framework (MACDF), designed to transform the traditional e-commerce search paradigm into a more user-centric, decision-focused process. MACDF leverages multiple specialized agents operating collaboratively, simulating professional shopping consultants, with the overarching goal of reducing user decision costs and enhancing the overall shopping experience.

\subsection{System Overview}
\begin{figure}[h]
  \centering
  \includegraphics[width=\linewidth]{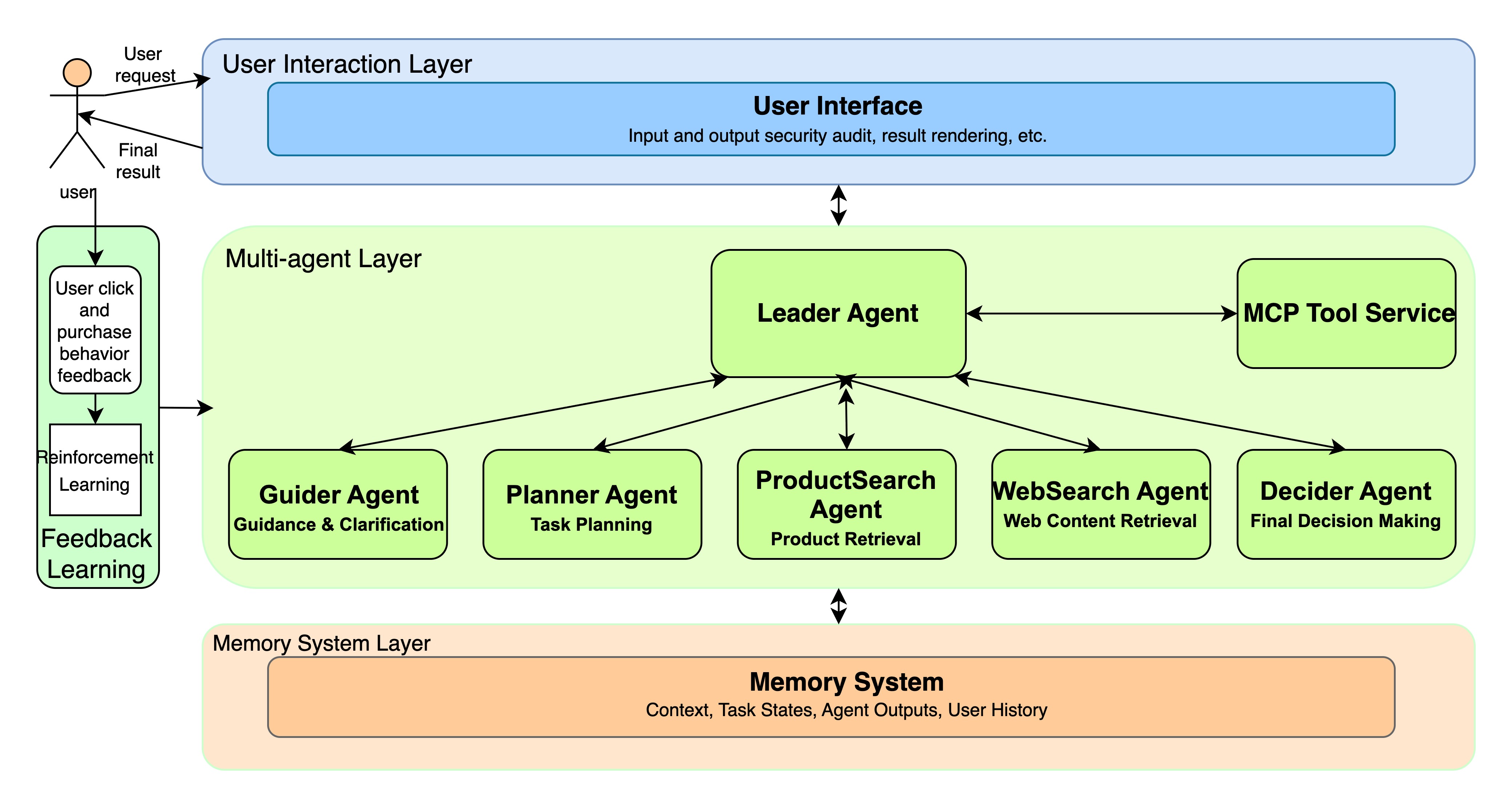}
  \caption{Multi-Agent Cognitive Decision Framework Overview}
  \label{figmain}
\end{figure}

Inspired by Simon's three-stage decision-making model\cite{simon1955behavioral}, the Multi-Agent Cognitive Decision Framework (MACDF) fundamentally reframes e-commerce search as a collaborative cognitive decision process, moving beyond the traditional retrieval-ranking pipeline. As illustrated in Figure ~\ref{figmain}, MACDF operates as a dynamic multi-agent collaboration system, where each agent plays a specific role in collectively optimizing the user's shopping journey.

The MACDF architecture comprises four key layers:
\begin{enumerate}
\item User Interaction Layer: Primarily responsible for user interfacing, including receiving user requests, security review of inputs and outputs, and result rendering.
\item  Multi-agent Layer: The core component of MACDF, consisting of several specialized SubAgents:
\begin{itemize}
    \item Leader Agent: Initially analyzes user requests. For ambiguous needs, it triggers the Guider Agent for clarification. For clear intents, it dispatches the Planner Agent to generate a Directed Acyclic Graph (DAG). It then orchestrates execution based on this graph, collects results into the Memory System, and finally schedules the Decider Agent to synthesize a final integrated result using the necessary outputs and context from memory. It also incorporates a reflection mechanism to autonomously evaluate results; if the outcome is deemed suboptimal, it re-dispatches the Planner for re-planning.
    \item Guider Agent: Provides clarification prompts when user needs are ambiguous to converge towards a clear intent. After a need is satisfied, it can suggest related queries to stimulate further engagement.
    \item Planner Agent: Decomposes and plans tasks based on the user's true intent, generating a task graph stored in the Memory System. Atomic tasks include product search, web search, or tool/service invocation.
    \item ProductSearch Agent: Executes product retrieval tasks generated by the Planner. It invokes product search APIs and semantic vector retrieval, then further enriches results by calling auxiliary agents (e.g., for reviews, detailed specifications).
    \item WebSearch Agent: Handles web search tasks by first generating search queries, then calling web search APIs to fetch relevant real-time external content.
    \item Decider Agent: Synthesizes the final decision outcome (e.g., purchasing advice, recommended products with justifications, follow-up questions) by integrating product results, web content, and relevant context from the Memory System.
\end{itemize}
\item Memory System Layer: Serves as the shared memory for all agents, storing and maintaining real-time contextual information, including SubAgent Context, Task States, Agent Outputs, and User History. Agents communicate and collaborate through this memory system.
\item Feedback Learning:The system can be further optimized based on user interactions (e.g., clicks, purchases) with MACDF, for instance, using reinforcement learning to enhance multi-agent collaboration strategies.
\end{enumerate}

\subsection{Leader Agent}
The Leader Agent first employs intent reasoning to determine whether to initiate an interactive clarification process via the Guider Agent, integrating two techniques:
\begin{enumerate}
    \item Demand Reasoning: Utilizes an LLM to reason over the user's input, generating a multi-intent distribution. Inputs include the current query, historical search/click behavior, context, relevant product attributes, and an intent classification space. The LLM directly outputs intent classifications and their confidence scores.
    \item Clarification Strategy: If the intent is judged ambiguous, the Guider Agent is activated to converge the fuzzy query towards a deterministic intent.
\end{enumerate}
Once the intent is clear, the Leader Agent delegates task planning to the Planner Agent for DAG generation. To ensure efficient execution of multiple tasks, a condition-driven coordination mechanism is designed, featuring key innovations:
\begin{enumerate}
    \item Conditional Scheduling Policy: The tasks are activated according to a condition:
    \begin{equation}
  t_k = \inf({t>t_{k-1}|\left\| e(t) \right\| > \delta*e^{-\alpha t}})
\end{equation}
In the formulated policy, the operator inf selects the infimum, which equates to the minimum value among all time instances t satisfying the specified condition. The notation $\left\| e(t) \right\|$ denotes the norm of the task state error vector. The variable $t\_k$ represents the timestamp of the k-th scheduling event. The parameter $\delta$ is the initial threshold parameter (set to 1.0 during the initial phase), while $\alpha$ is the decay coefficient that controls the exponential decay rate of the threshold over time, thereby reflecting the increasing urgency of the task. Scheduling is triggered exclusively when the change in a subtask's readiness state exceeds this adaptive threshold.

     \item Cross-Agent Memory Sharing: Facilitates knowledge transfer between heterogeneous agents via the global Memory System.
\end{enumerate}
After collecting subtask outputs and updating memory, the Leader schedules the Decider Agent for result integration. Incorporating a reflection design inspired by ReAct\cite{yao2023react}, the Leader Agent employs a two-stage evaluator:

\begin{enumerate}
    \item Real-time Reflection: Uses online metrics for relevance and predicted user satisfaction:
    \begin{equation}
        score = \gamma * Relevance + (1-\gamma) * User\_Satisfaction
    \end{equation}
    where $Relevance$ is the relevance score between user's intention and candidate items, and $User\_Satisfaction$ is user's satisfaction score of final result, both of which are predicted by Leader. If the score falls below a predefined threshold, the result is deemed suboptimal, and shall trigger the Planner for re-planning.
    \item Long-term Adaptation: Records failure cases into a reflection knowledge base, driving iterative policy updates and enabling closed-loop continuous improvement.
\end{enumerate}
\subsection{Guider Agent}
The Guider Agent acts as a dynamic cognitive coordinator, aiming to transform ambiguous user queries into actionable tasks and proactively stimulate potential needs post-satisfaction. Drawing from cognitive psychology on cognitive convergence\cite{roschelle1995construction}, it employs a "clarification-guidance-context synergy" mechanism. Its clarification engine uses a hierarchical, brain-inspired reasoning process\cite{wang2025hierarchical}: first identifying entities and constraints via LLM function calling for semantic slot filling; then, incorporating user feedback, it utilizes a Q-learning\cite{kostrikov2021offline} based reward function: 
\begin{equation}
    R(s,a) = \alpha * Intent\_Gain - \beta * Interaction\_Rounds
\end{equation}
Here, $s$ denotes the current dialogue state of the Guider, encompassing the user’s intent, the distribution of filled semantic slots, dialogue history, and user behavior profile. $a$ represents the next guiding action to be taken by the Guider in state $s$. $Intent\_Gain$ refers to the gain in intent confidence—a reward signal quantifying how much the system’s understanding of the user’s intent improves as a result of executing action a and receiving user feedback. This metric evaluates the effectiveness and utility of the Guider Agent’s inquiry. The variable $Interaction\_Rounds$ counts the number of interaction turns, while $\alpha$ and $\beta$ are hyperparameters that balance the two competing objectives in the reward function. The goal of this reward function is to facilitate rapid convergence of the user’s vague intent into a clear and executable task description by posing the most critical and incisive questions in as few interaction rounds as possible.

Beyond clarification, the Guider Agent possesses post-reply guidance capability, using the LLM to generate potential follow-up queries based on user history and current results (e.g., suggesting "hiking shoes" after recommending "trekking poles"). It operates in synergy with the global memory module, following principles analogous to hippocampal-neocortical interaction\cite{mcclelland1995there} for state synchronization and co-activation with other agents.
\subsection{Planner Agent}
The Planner Agent functions as the core reasoning engine, translating user intent into an executable task graph $G=(V, E)$, where vertices $V$ represent atomic tasks and edges $E$ represent dependencies. Planning involves two steps:
\begin{enumerate}
    \item Semantic Parsing \& Atomic Task Generation: Identifies intents and decomposes them into four atomic task types: ProductSearch, WebSearch, Tool Call, and Decision.
    \item Dependency Construction: Automatically generates edges E based on task semantic constraints (e.g., Task\_4 depends on outputs from Task\_1, Task\_2, Task\_3). The graph is stored in the Memory System's graph database.
\end{enumerate}
Through the above steps, the Planner can extend simple queries into multiple related ProductSearch subtask queries based on user intent, decompose complex queries into simpler subtasks, generate WebSearch subtasks for information requiring internet retrieval, produce tool-oriented subtasks based on MCP\cite{anthropic2024introducing} for functional needs, and additionally create decision-making tasks.
A dual-constraint mechanism ensures feasibility:
\begin{enumerate}
    \item Hard Constraints: Symbolic logic enforces inviolable rules (e.g., product comparison requires $\geq$2 product search outputs; time-sensitive queries mandate web search; conflict triggers re-planning).
    \item  Soft Constraints: Incorporated as penalty terms in a reward function (e.g., prioritizing merged API calls; prioritizing sensitive tasks like price comparison).
\end{enumerate}
\subsection{ProductSearch Agent}
This agent translates structured retrieval tasks into candidate product lists. Its innovation lies in a cognitively-enhanced search paradigm combining hybrid retrieval, information enrichment, and LLM-based fusion ranking.
It employs a dual-channel hybrid retrieval mechanism (keyword-based API calls and semantic vector similarity using contrastively learned embeddings). Post-retrieval, an enrichment module collaborates with auxiliary agents to aggregate attributes and generate concise summaries (e.g., pros/cons). Finally, an LLM performs fusion ranking using Chain-of-Thought\cite{wei2022chain} prompts for need-product alignment, comparative reasoning using multi-source evidence, and personalized calibration based on user history.
\subsection{WebSearch Agent}
This agent addresses the limitations of traditional e-commerce search regarding real-time and external information. Its "Retrieve-Filter-Integrate" three-tier architecture enables a shift from passive retrieval to active cognitive extension.
The Web Retriever uses the LLM to generate and expand search queries based on the current context, fetching diverse real-time results via search APIs. A filtering module then scores results using:
\begin{equation}
    score = \alpha*relevance + \beta*authority + \gamma*freshness
\end{equation}
where weights $\alpha$, $\beta$, $\gamma$ balance importance, and applies thresholds for efficiency. Finally, an integration stage uses the LLM to semantically extract and compress filtered content into high-quality summaries relevant to the user's request, providing crucial cross-domain information for downstream decision-making.
\subsection{Decider Agent}
The Decider Agent serves as the final decision-making hub of the multi-agent cognitive decision-making framework, responsible for integrating multi-source information and generating interpretable and executable shopping recommendations. Its design integrates Simon's three-stage decision-making model\cite{simon1955behavioral} and multi-criteria decision-making theory\cite{keeney1993decisions}, achieving a paradigm shift from "passive retrieval" to "active expert decision-making."

First, it aggregates multi-source contextual information from the memory system, including: product information and review summaries from the ProductSearch Agent, real-time online information from the WebSearch Agent, user requirement clarification results from the Guider Agent, as well as user historical preferences and budget constraints, to construct a unified decision-making context.

Second, it performs decision analysis based on the aggregated information. To balance different evaluation dimensions, it references the Analytic Hierarchy Process (AHP)\cite{vaidya2006analytic} and Pareto Optimality\cite{censor1977pareto}, combining real-time context to form a dynamic multi-criteria evaluation matrix. This matrix comprehensively considers:
\begin{itemize}
    \item Functional performance: core functions, quality, and consumer evaluations of products;
    \item Economic cost: the relationship between price and the user's budget;
    \item Risk reliability: assessment of product reliability and after-sales support based on user reviews and external online search content;
    \item Preferences and constraints: integration of user historical preferences and specific constraints provided by the Planner Agent.
\end{itemize}

Finally, it recommends products and generates reasoning. Based on the optimal option derived from the multi-criteria analysis, it invokes the large language model through a specifically designed decision prompt that integrates multi-source signals. The output includes:
\begin{itemize}
    \item Purchasing advice: tailored to the user's personalized needs;
    \item Product recommendation: presenting the optimal product and its key features;
    \item Reasoning for purchase: generating explainable justifications based on functionality, cost-effectiveness, and reliability;
    \item  Follow-up question: stimulating further needs to refine the decision process.
\end{itemize}

To address real-time dynamic changes in user needs, the Decider possesses self-adaptive capabilities:
\begin{itemize}
    \item Self-evaluation and feedback: After generating product results, the Decider evaluates its own decision quality based on the user's immediate feedback. If the evaluation is unsatisfactory or the user raises new needs, it sends a signal to the Leader to trigger the Planner for new task planning or search strategy adjustment.
    \item Model update and memory: When the system frequently encounters similar needs and demonstrates stable decision performance, the Decider can store experience samples in the memory system for continuously updating its decision model, improving future recommendation accuracy and timeliness.
\end{itemize}

\section{Experiments}
To comprehensively evaluate the effectiveness of our proposed Multi-Agent Cognitive Decision Framework (MACDF), we conducted an extensive set of experiments, ranging from offline evaluations to online testing.
\subsection{Experimental Setup}
\subsubsection{E-Commerce Cognitive Decision Benchmark}
The current lack of a standardized benchmark for assessing cognitive decision-making capabilities in e-commerce search motivated us to construct a new evaluation dataset: the E-Commerce Cognitive Decision Benchmark (ECCD-Bench). This dataset is derived from desensitized user queries and their session contexts sourced from JD real-world e-commerce platform. Expert annotators manually labeled these instances based on their genuine user intents and the ideal system responses. ECCD-Bench comprises 10k queries across three distinct types, designed to test different system capabilities:
\begin{itemize}
    \item \textbf{Simple Queries} (30\%):Direct requests for specific products or categories (e.g., "iPhone 15").
    \item \textbf{Complex Queries} (50\%):Queries requiring advanced cognitive skills, further subdivided into:
    \begin{itemize}
        \item \textbf{Reasoning Queries}:Require inference based on world knowledge (e.g., "housewarming gift for a friend").
        \item \textbf{Multi-Constraint Queries}:Involve multiple specific filtering conditions (e.g., "wireless headphones under 500 RMB with active noise cancellation and 30-hour battery life").
        \item \textbf{Negation Intent}: Explicitly exclude certain items (e.g., "recommend a smartphone around 1000 RMB but not Redmi").
        \item \textbf{Ambiguous Intent}: Contain unclear demands, such as "gift for someone".
    \end{itemize}
    \item \textbf{Consultative Queries} (20\%): Focus on seeking advice or comparisons rather than direct retrieval:
    \begin{itemize}
        \item \textbf{Product Q\&A}:Ask about product features or seek recommendations (e.g., "What is the screen size of Huawei Mate70 pro?").
        \item \textbf{Product Comparison}:Request comparisons between different products (e.g., "Compare iPhone 16 and Mate70").
        \item \textbf{General Q\&A}:General Q\&A with potential shopping needs (e.g., "What is the normal heart rate for an adult?").
    \end{itemize}
\end{itemize}
\subsubsection{Baseline Systems}
We compare MACDF against two strong baseline systems, representing state-of-the-art traditional paradigms and a powerful direct RAG approach, respectively:
\begin{itemize}
    \item Online System:This baseline represents our current production system's main pipeline, which primarily relies on a retrieve-and-rank paradigm. LLMs are extensively used to enhance both the retrieval and ranking stages within the overall search system.
    \item RAG Method:As a controlled comparison, this system injects retrieved candidate products and their metadata into a prompt using a RAG approach. The LLM's task here is to re-rank the product results and generate justification text for recommendations. This baseline tests whether simply grafting an LLM onto the old paradigm is sufficiently effective.
\end{itemize}
\subsubsection{Evaluation Metrics}
We employ a multi-faceted evaluation strategy:
\begin{itemize}
    \item \textbf{Offline Evaluation (on ECCD-Bench)}:
    \begin{itemize}
        \item \textbf{Accuracy@K  (ACC@K)}:Measures the ability to retrieve relevant items matching user needs within the top K results. Reflecting typical online result displays, we report results for K=5.
        \item \textbf{User Demand Satisfaction (UDS)}:A human evaluation metric scored on a 1-3 scale (corresponding to Not Satisfied, Partially Satisfied, and Fully Satisfied). Expert annotators score the system's final response (including recommended products, answer content, key selling points, and guidance) based on its completeness, correctness, and usefulness in addressing both explicit and implicit user needs.
    \end{itemize}
    \item \textbf{Online Evaluation (via A/B Testing)}:
    \begin{itemize}
        \item \textbf{User Click-Through Rate (UCTR)}:the average number of product clicks per searching user.
        \begin{displaymath}
            UCTR = \frac{num\_of\_click}{UV_{search}}
        \end{displaymath}
        \item \textbf{User Conversion Rate (UCVR)}: the average number of orders placed per searching user.
        \begin{displaymath}
            UCVR = \frac{num\_of\_order}{UV_{search}}
        \end{displaymath}
        \item \textbf{Average Reformulation Count (ARC)}:The average number of times a user needs to modify their initial query within a session. A lower ARC indicates the system's superior ability to understand user intent early, thereby reducing user decision cost.
    \end{itemize}
\end{itemize}
\subsubsection{Implementation Details}
The core implementations of the various agent components within the MACDF framework are as follows:
\begin{itemize}
    \item \textbf{Leader Agent}:Utilizes Deepseek-V3\cite{liu2024deepseek} as the base model, with capabilities primarily optimized for multi-agent coordination and scheduling via prompt engineering.
    \item \textbf{Guider Agent and Planner Agent}: Utilizes Qwen3-8B\cite{yang2025qwen3} as the base model, with guidance and planning capabilities enhanced by distilling knowledge from Deepseek-R1\cite{guo2025deepseek}.
    \item \textbf{ProductSearch Agent}:Combines semantic vector retrieval with traditional retrieval methods to recall a candidate set of TOP100 products.
    \item \textbf{WebSearch Agent}:Employs the publicly available ZhiPu Search API for real-time web information acquisition, with integration and generation handled by a Qwen3-8B[32] model.
    \item \textbf{Decider Agent}:Utilizes Deepseek-V3 as the base model to generate comprehensive decision suggestions.
\end{itemize}
LLM training employed a proprietary framework on Huawei Ascend 910B clusters, using the AdamW optimizer (lr=2e-6). A/B tests deployed MACDF to a 1\% traffic bucket on JD search platform for two weeks.
\subsection{Main Results}
\subsubsection{Offline Evaluation on ECCD-Bench}
The offline evaluation results on the ECCD-Bench dataset are summarized in Table~\ref{tab:acc} and Table~\ref{tab:uds} , which reports the proportion of products satisfying the user’s intent among the top-5 recommendations, measured by accuracy@5 and UDS@5.

\begin{table*}
  \caption{Offline Evaluation Accuracy Results on ECCD-Bench}
  \label{tab:acc}
  \begin{tabular}{cccc}
    \toprule
    Metric: Accuracy@top5 & online system(base) & RAG method & MACDF \\
    \midrule
    Simple Queries & \textbf{0.87} & 0.83 & \textbf{0.87} \\
    Negation Intent & 0.07 & 0.12 & \textbf{0.86} \\
    General Q\&A & 0.48 & 0.48 & \textbf{0.82} \\
    Reasoning Queries & 0.66 & 0.66 & \textbf{0.75} \\
    Multi-Constraint Queries & 0.41 & 0.41 & \textbf{0.80} \\
    Ambiguous Intent & 0.81 & 0.80 & \textbf{0.92} \\
    Product Comparison & 0.70 & 0.62 & \textbf{0.76}  \\
     Product Q\&A & 0.50 & 0.51 & \textbf{0.55}\\
    \bottomrule
  \end{tabular}
\end{table*}

\begin{table*}
  \caption{Offline Evaluation UDS Results on ECCD-Bench}
  \label{tab:uds}
  \begin{tabular}{cccc}
    \toprule
    Metric: UDS@top5 & online system(base) & RAG method & MACDF \\
    \midrule
    Simple Queries & 2.64 & 2.55 & \textbf{2.70} \\
    Negation Intent & 1.17 & 1.23 & \textbf{2.76} \\
    General Q\&A & 1.63 & 2.21 & \textbf{2.92} \\
    Reasoning Queries & 2.25 & 2.17 & \textbf{2.72} \\
    Multi-Constraint Queries & 1.83 & 1.82 & \textbf{2.76} \\
    Ambiguous Intent & 2.74 & 2.73 & \textbf{2.80} \\
    Product Comparison & 1.80 & 2.57 & \textbf{2.94}  \\
     Product Q\&A & 1.81 & 2.51 & \textbf{2.83}\\
    \bottomrule
  \end{tabular}
\end{table*}
The experimental results demonstrate that, in terms of product accuracy metrics, the proposed MACDF exhibits a significant advantage over the conventional online system and the RAG baseline when handling complex query scenarios. Particularly in challenging tasks such as negative intent, general QA, multi-constraint intent, and reasoning intent, MACDF achieves substantial improvements in Accuracy@Top5. This indicates that MACDF can deeply comprehend implicit user needs and complex constraints, effectively emulating a professional shopping assistant by dynamically integrating information and inferencing user requirements. As a result, it achieves notable breakthroughs in complex scenarios and successfully mitigates the semantic gap inherent in traditional paradigms.

Furthermore, MACDF shows comprehensive improvements in the user demand satisfaction (UDS) metric, which reflects the overall user experience. Especially in intents such as general Q\&A, comparative queries, and product inquiries, users are able to obtain satisfactory results without the need for multiple clicks or cross-platform searches. This underscores MACDF’s capability to provide professional, assistant-like decision support—going beyond mere product matching optimization.

In contrast, although the RAG baseline exhibits certain improvements in some aspects, its overall performance remains inferior to that of MACDF. These findings validate that the multi-agent collaborative framework is more effective than relying solely on retrieval-augmented generation.
\subsubsection{Online A/B Test Results}
In our online A/B test, a certain proportion of users in the small-traffic bucket exited before results were returned due to the higher latency of MACDF. To ensure a fair comparison, we excluded these requests from the metric calculations. Furthermore, we only compare the top10 results between online base system and MACDF. The experimental results, presented in terms of relative percentage changes, are shown in Table~\ref{tab:online}.
\begin{table}
  \caption{Online A/B Test Results (Relative Change \%)}
  \label{tab:online}
  \begin{tabular}{cccc}
    \toprule
    system & UCTR & UCVR & ARC\\
    \midrule
    online base & 0.0\% & 0.0\% & 0.0\% \\
    MACDF & +3.9\%(P=0.037)& +6.5\%(P=0.025) &-7.7\%(P=0.004)\\
  \bottomrule
\end{tabular}
\end{table}

Through a small-scale online experiment, the proposed MACDF framework demonstrated significant improvements over the baseline system. Specifically, UCVR increased by 6.5\%, indicating that MACDF facilitates more purchase behaviors by offering more professional and tailored decision support. UCTR rose by 3.9\%, suggesting that the result list generated by MACDF is not only more relevant but also accompanied by compelling justifications and guided content that better stimulates users’ click-through interest, reflecting greater user trust and willingness for deeper exploration. ARC decreased by 7.7\%, implying that users no longer need to repeatedly refine their queries to “teach” the system their intent, as MACDF delivers satisfactory results in the first interaction.

We attribute these gains to MACDF’s enhanced ability to comprehend user intent (leading to lower ARC) → present more engaging and relevant results (driving higher UCTR) → and ultimately deliver information sufficient to support purchase decisions (resulting in higher UCVR). This cycle underscores how "cognitive decision optimization" outperforms "mere product matching".

Drill-down analysis further reveals that under MACDF, the proportion of complex user intents and question-answering intents increased substantially, which constitutes the primary source of the observed gains. The experiment also identified an area for improvement: system latency. A number of users exited before results were returned, indicating that the current framework incurs non-trivial computational overhead.

\subsection{Computational Efficiency Evaluation}
Despite the MACDF framework's superior performance on effectiveness metrics, it is essential to evaluate whether its computational efficiency meets practical application requirements. We tested the response latency of each method on servers equipped with Huawei Ascend 910 AI processors.

MACDF employs a streaming output mechanism. The average time-to-first-token is 1.8 seconds, and the average total response time for a complete output—which typically includes a 100–200 word reply, 10 recommended products, corresponding justification, and follow-up guidance—is around 15 seconds. This is higher than the current online baseline system, which achieves an average latency of 0.8 seconds with non-streaming output. The extended latency of MACDF is primarily attributable to its 
sophisticated multi-agent collaboration and decision-making processes.

While the extended latency in simple-intent scenarios may noticeably impact user experience, a first-token delay of 1.8 seconds remains acceptable for complex e-commerce search tasks—particularly given MACDF’s ability to significantly reduce the user’s overall decision-making time.

To mitigate latency, we have implemented performance optimizations including asynchronous execution and an intelligent caching mechanism. For instance, external web search and internal product search operations are executed in parallel, and frequently encountered queries are cached to avoid redundant computations, thereby reducing both time-to-first-token and end-to-end response time.
\section{Discussion}
\subsection{Discussion of Results}
The experimental results demonstrate that the MACDF framework significantly outperforms the traditional retrieval-ranking paradigm across multiple dimensions, validating our core hypothesis: reframing e-commerce search as a multi-agent cognitive decision-making process effectively bridges the gap between conventional systems and the actual user decision-making process.

Compared to traditional search systems and LLM-based RAG systems, MACDF exhibits superior performance in decision efficiency and complex query handling. This indicates that mere algorithmic optimizations or joint multi-scenario enhancements are insufficient to address the fundamental challenges—instead, a paradigm shift at the architectural level is necessary.
\subsection{Practical Application Value}
MACDF holds not only academic novelty but also substantial practical potential. After being deployed on JD e-commerce platform, small-scale A/B testing metrics confirmed that MACDF can improve conversion rates, leading to direct revenue growth, while also reducing users’ decision-making cost and enhancing satisfaction—thereby mitigating user churn to external platforms.

Furthermore, the rich user interaction data generated by MACDF provides valuable resources for subsequent optimization: decision process logs enable deeper insight into user decision mechanisms, agent collaboration data guides the improvement of multi-agent systems, and user feedback supports continuous reinforcement learning.
\subsection{Limitations and Future Work}
Despite its strong performance, the MACDF framework has certain limitations. First, its computational demand is considerably higher, with an average query processing time approximately 18.75× longer than that of traditional methods (15 seconds vs. 0.8 seconds). Although the total decision time is reduced, real-time responsiveness remains a challenge. Future research may explore model compression, pre-computation, and caching strategies to improve response speed.

Second, the current evaluation of MACDF primarily relies on online experiments and short-term metrics. The long-term impact on user retention and customer lifetime value requires further investigation. Future work should establish a long-term evaluation framework to analyze MACDF’s influence on user loyalty and repurchase behavior.Finally, the performance improvement of MACDF stems from the collaborative synergy of multiple agents and the integrated effect of various algorithms and strategies. Further systematic ablation studies are required to determine which collaborative combination is the most critical.
\section{Conclusion}

This study addresses the fundamental misalignment between traditional e-commerce search paradigms and users’ cognitive decision-making processes by proposing the Multi-Agent Cognitive Decision Framework (MACDF). MACDF reframes e-commerce search as a collaborative cognitive decision process, utilizing multi-agent collaboration to explicitly optimize users' decision costs.

Experimental results demonstrate that MACDF significantly outperforms existing baseline methods on both traditional e-commerce metrics (UCTR, UCVR) and decision-cost metrics (Average Reformulation Count, ARC). This validates the effectiveness of MACDF and confirms that optimizing the decision process itself can simultaneously enhance both user experience and business value.

The theoretical contribution of MACDF lies in introducing a new paradigm for e-commerce search research—shifting from an information retrieval perspective to a cognitive decision perspective. Its practical contribution is the design of a feasible multi-agent architecture that addresses key limitations in traditional e-commerce search, including the semantic gap, high decision cost, and lack of professional guidance.

Future work will focus on three directions: (1) improving the computational efficiency of MACDF to reduce response latency; (2) establishing a long-term evaluation framework to analyze the impact of MACDF on user lifetime value; (3) identifying the most critical agents and optimal strategy combinations through systematic ablation studies.

\bibliographystyle{ACM-Reference-Format}
\bibliography{sample-base}

\end{document}